\documentclass[11pt]{article}
\usepackage{amsmath}
\usepackage{graphicx}
\usepackage{ textcomp }
\usepackage{ amssymb }
\usepackage{float}
\graphicspath{ {images/} }
\usepackage{booktabs} 
\usepackage[table]{xcolor} 
\usepackage{authblk}

\usepackage{tabularx} 
\usepackage{rotating} 
\usepackage{siunitx} 
\usepackage{caption}

\usepackage{hyperref}
\usepackage[numbers,sort&compress]{natbib}

\usepackage{adjustbox}
\usepackage[margin=0.5in]{geometry}
\begin{document}

\makeatletter
\newcommand*{\rom}[1]{\expandafter\@slowromancap\romannumeral #1@}
\makeatother

\title{In-Context Learning can distort the relationship \\between sequence likelihoods and biological fitness}

\author{
  Pranav Kantroo\textsuperscript{1,6,}\footnote{pranav.kantroo@yale.edu} ,
  G\"unter P. Wagner\textsuperscript{2,3,4}, 
  Benjamin B. Machta\textsuperscript{5,6,}\footnote{benjamin.machta@yale.edu}
  \\
  \small
  \textsuperscript{1}Computational Biology and Bioinformatics Program, Yale University, New Haven, CT-06520, USA \\
  \textsuperscript{2}Emeritus, Department of Ecology and Evolutionary Biology, Yale University, New Haven, CT-06520, USA \\
  \textsuperscript{3}Department of Evolutionary Biology, University of Vienna, Djerassi Platz 1, A-1030 Vienna, Austria \\
  \textsuperscript{4}Hagler Institute for Advanced Studies, Texas A\&M, College Station, TX-77843, USA \\
  \textsuperscript{5}Department of Physics, Yale University, New Haven, CT-06520, USA \\
  \textsuperscript{6}Quantitative Biology Institute, Yale University, New Haven, CT-06520, USA 

}
\date{}
\maketitle
\vspace{-3em}

\begin{abstract}
\noindent Language models have emerged as powerful predictors of the viability of biological sequences. During training these models learn the rules of the grammar obeyed by sequences of amino acids or nucleotides. Once trained, these models can take a sequence as input and produce a likelihood score as an output; a higher likelihood implies adherence to the learned grammar and correlates with experimental fitness measurements. Here we show that in-context learning can distort the relationship between fitness and likelihood scores of sequences. This phenomenon most prominently manifests as anomalously high likelihood scores for sequences that contain repeated motifs. We use protein language models with different architectures trained on the masked language modeling objective for our experiments, and find transformer-based models to be particularly vulnerable to this effect. This behavior is mediated by a look-up operation where the model seeks the identity of the masked position by using the other copy of the repeated motif as a reference. This retrieval behavior can override the model's learned priors. This phenomenon persists for imperfectly repeated sequences, and extends to other kinds of biologically relevant features such as reversed complement motifs in RNA sequences that fold into hairpin structures. 
\end{abstract}

\section{Introduction}

Language models have found widespread use in biological settings \citep{lin2023evolutionary, zhou2023dnabert2, penic2024rinalmo, theodoris2023transfer, trauble2025multi}. Their success in facilitating protein design and fitness prediction~\citep{lin2023evolutionary, nijkamp2023progen2} has prompted their application in other domains like DNA \citep{zhou2023dnabert2, nguyen2024sequence} and RNA-based \citep{penic2024rinalmo, shulgina2024rna} sequences. A significant aspect of their appeal lies in their ability to assess functional plausibility -- the higher the likelihood score of a sequence relative to some related variant, the higher its experimentally measured fitness tends to be \citep{ruffolo2024designing, notin2024proteingym, benegas2025benchmarking}.

Although the statistical connection between fitness and likelihood scores has been validated for proteins, DNA and RNA~\citep{notin2024proteingym, benegas2025benchmarking, brixi2025genome, shulgina2024rna}, it comes with important caveats~\citep{ding2024protein, shaw2023removing, gordon2024protein}. For instance, this relationship can be complicated by the biases that the model picks up over its training. Protein language models have been shown to exhibit clade-specific biases wherein proteins from some species can have systematically higher likelihoods compared to their variants derived from other species~\citep{ding2024protein}. This has been traced to compositional imbalances within the dataset that was used to train the model. Such biases can undermine design workflows that use likelihood scores to guide the exploration of the sequence landscape. Identifying the factors that contribute to such distortions in the fitness-likelihood relationship can help us develop workarounds for downstream applications~\citep{shaw2023removing, gordon2024protein}. 

\begin{figure}[!]
    \centering
    \begin{minipage}{0.61\textwidth} 
        \centering
        \includegraphics[width=\textwidth]{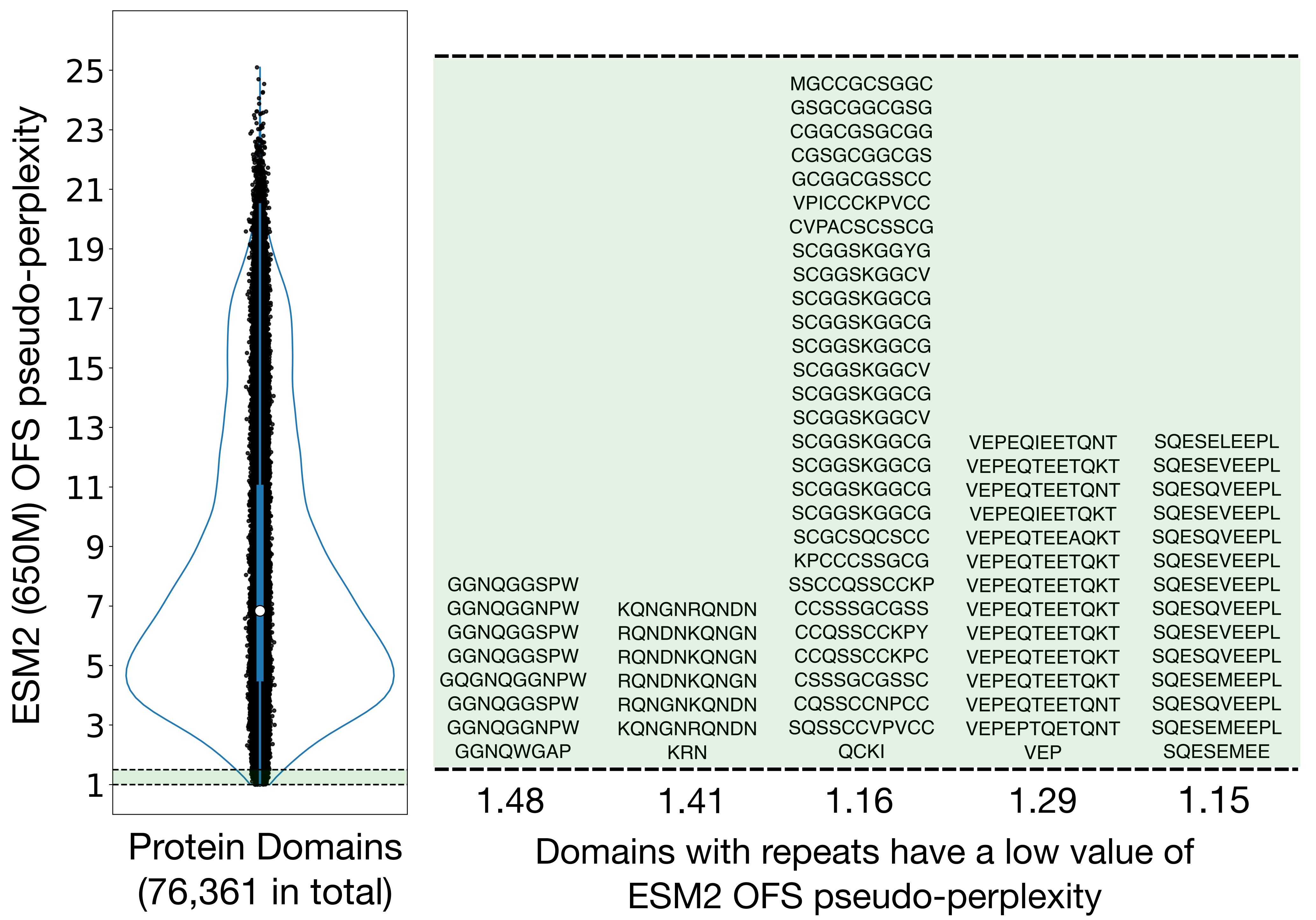}
    \end{minipage}
    \hfill
    \begin{minipage}{0.38\textwidth} 
        \captionof{figure}{\small \textbf{Proteins with repeated motifs have strikingly low pseudo-perplexity scores} A visualization of the ESM2 OFS pseudo-perplexity distribution of protein domains parsed from a diverse set of protein sequences. We manually scanned through the set of domains that have low pseudo-perplexity scores (left panel: values from 1 to 1.5 shaded in green in the plot). We found that a significant proportion of these low pseudo-perplexity sequences have repeated motifs in them. We have visualized the sequences of five such protein domains in the right panel (green shaded box). The text for each of these five sequences has been wrapped in a way to make the repetitive structure apparent.}
        \label{fig:schematic}
    \end{minipage}
\end{figure}

Towards this end, we sought to identify commonalities across protein sequences with the highest pseudo-likelihood scores~\citep{salazar2019masked}, as evaluated by the masked language model ESM2~\citep{lin2023evolutionary}. We parsed a large set of diverse protein sequences into their constituent domains~\citep{marchler2004cd, lu2020cdd} (see Methods for details) and calculated their pseudo-perplexity through the One Fell Swoop (OFS) approach~\citep{kantroo2024pseudo} -- a low value of pseudo-perplexity maps to a high pseudo-likelihood score. We visually inspected the set of sequences with low pseudo-perplexity scores and found that a significant proportion of such sequences are composed of repeated motifs (Figure~\ref{fig:schematic}). We hypothesized that the reason for the strikingly low pseudo-perplexity scores of such sequences is that the model uses other existing copies of the repeated motif to look up the identity of the residue at the masked position. In this work, we show that this look-up operation, a form of in-context learning~\citep{reddy2023mechanistic, chan2022data, singh2023transient, elhage2021mathematical, olsson2022context}, is what leads to the anomalous scores that we observe for these sequences. 

Learning in the traditional sense involves minimizing the loss for some given training objective through explicit updates to a model's weights over iterative steps. Large language models however have been shown to exhibit an emergent form of learning where they learn from the presented context \textit{as is} without any updates to the weights themselves~\citep{brown2020language, chan2022data, akyurek2022learning}. This \textit{in-context} learning process happens through transformations of the input as it passes through the layers of the model~\citep{von2023transformers}. This capability allows the model to learn from the provided context at the time of inference and dynamically adjust its predictions, bypassing the need for any task-specific training~\citep{brown2020language}. Given its importance and utility, this feature has been extensively studied within autoregressive transformers trained on natural language as well as synthetic datasets~\citep{brown2020language, chan2022data, park2024competition, akyurek2022learning}. Here we explore the ramifications of this behavior as it pertains to masked language models trained on biological sequences, and connect our findings with observations made in the natural language context. 

\section{Results}

\subsection{Repetition can induce an uncertainty collapse in masked language models}

\begin{figure}[!]
    \centering
    \includegraphics[width=1\textwidth]{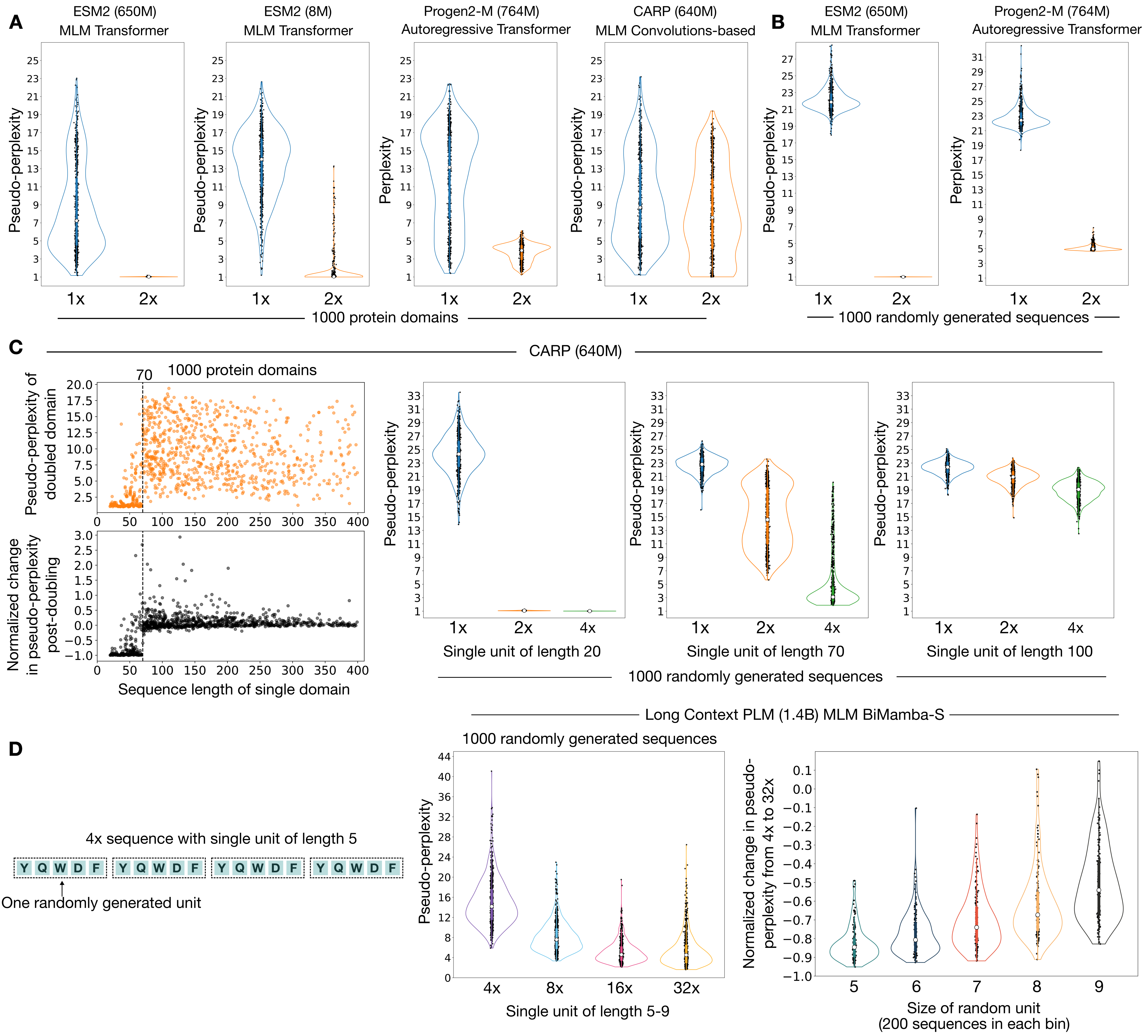}
    \caption{\small \textbf{Repetition can induce an uncertainty collapse in masked language models}                  Pseudo-perplexity is a measure of the model's uncertainty in its predictions with a value of 1 signifying complete and absolute certainty.
                            (A) We show violin plots of the distribution of pseudo-perplexity scores for 1000 protein domains. The 1x distribution in blue denotes natural protein sequences and the 2x distribution in orange denotes doubled sequences. A doubled sequence is generated by appending a copy of a sequence to itself. Transformer-based masked language models (ESM2) exhibit an uncertainty collapse when presented with doubled sequences -- the pseudo-perplexity of the doubled sequences is approximately one, the lowest value that it can take. Progen-M, an autoregressive transformer-based model exhibits a significant decline in its perplexity for doubled sequences. However, the convolutions-based masked language model (CARP) exhibits markedly different behavior.
                            (B) Randomly generated sequences with the same lengths as natural protein domains also exhibit an uncertainty collapse after being doubled in ESM2. Likewise, we observe a sharp decline in the perplexity of these doubled random sequences in Progen-M.
                            (C) CARP (640M) exhibits uncertainty collapse for protein sequences shorter than approximately 70 residues.
                            It also exhibits an uncertainty collapse for randomly generated repeating units of size 20. However, we only observe a slow progressive decline for repeating units of size 100 with an increase in the multiplicity. The model is particularly susceptible to an increase in the multiplicity of repeating units of size 70.
                            (D) LC-PLM (1.4B), a BiMamba-S based protein language model does not exhibit uncertainty collapse, even for short repeating units that vary between 5-9 residues at multiplicities as high as 32x. It only shows a progressive decline in uncertainty with increasing multiplicity of the repeats. The decline in pseudo-perplexity from 4x to 32x units is most pronounced for the shortest repeating unit of 5 residues. The effect gets weaker as the size of the repeating unit increases.}
    \label{fig:Mirage}
\end{figure}

The observation about low ESM2 OFS pseudo-perplexity values of sequences with repeated motifs (Figure~\ref{fig:schematic}) drove us to look into how \textit{doubling} a sequence -- appending a copy of the sequence to itself, affects the model's assessment. In order to rule out the possibility that our observations may be a quirk of the One Fell Swoop calculation (OFS)~\citep{kantroo2024pseudo}, as opposed to an inherent property of the model, we compute pseudo-perplexity by masking the residues of a sequence one at a time for this experiment~\citep{salazar2019masked}. 

We first use ESM2 (650M)~\citep{lin2023evolutionary}, a transformer-based~\citep{vaswani2017attention} model with 650 million parameters, trained on the masked language modeling objective~\citep{devlin2018bert}. We find that doubling a protein domain leads to a complete collapse in the model's uncertainty -- the pseudo-perplexity of a doubled sequence is approximately one, the lowest value that it can take (Figure~\ref{fig:Mirage}:A). The much smaller ESM2 (8M) also exhibits this uncertainty collapse indicating that this behavior is not exclusive to larger-scale models (Figure~\ref{fig:Mirage}:A). We repeat this experiment for Progen2-M~\citep{nijkamp2023progen2}, an autoregressive transformer-based model~\citep{radford2019language}, and observe qualitatively similar behavior -- the perplexity of the sequences drops significantly after being doubled (Figure~\ref{fig:Mirage}:A). The reduction in perplexity, as opposed to a complete collapse, is explained by the fact that autoregressive models can only access the preceding sequence context in making their predictions, unlike masked language models that utilize bidirectional context. We repeated this doubling experiment with randomly generated amino acid sequences that have the same lengths as the natural protein domains. The change in the distribution of the statistics of the sequences is reflected in their significantly up-shifted pseudo-perplexity and perplexity distribution (Figure~\ref{fig:Mirage}:B). Even so, doubling these nonsensical sequences triggers an uncertainty collapse in ESM2 and a significant drop in perplexity for Progen-M (Figure~\ref{fig:Mirage}:B). 

Next, we ran this experiment for the convolutions-based protein language model CARP (640M), that is trained on the masked language modeling objective, and has been shown to be competitive with transformer-based models on downstream applications~\citep{yang2024convolutions}. We observe that this model exhibits markedly different behavior (Figure~\ref{fig:Mirage}:C). A closer look at the data reveals that although the model does exhibit an uncertainty collapse, it does so only for repeating units that are shorter than approximately 70 residues (Figure~\ref{fig:Mirage}:C). This means that the model can easily recognize two repeating units that are smaller than 70 residues each, but is unable to do so confidently if they are larger than this threshold -- effectively defining the operational memory of the model. We then repeated this experiment with randomly generated sequences, but used repeating units with a fixed size of 20, 70 and 100 residues, with a multiplicity of 1x, 2x and 4x units. We observe that repeating units of size 20 exhibit a complete uncertainty collapse, while repeating units of size 100 only exhibit a progressive decline in uncertainty with increasing multiplicity (Figure~\ref{fig:Mirage}:C). Repeating units of size 70, that lie at the precipice of the transition, exhibit a heightened sensitivity to the multiplicity of the repeats (Figure~\ref{fig:Mirage}:C). These observations indicate that there are two distinct regimes marked by an uncertainty collapse and progressive decline in CARP (640M) that are separated by a transition between these two behaviors.

Finally, we experimented with the long-context protein language model -- LC-PLM (1.4B), that is based on the BiMamba-S architecture and trained on the masked language modeling objective~\citep{wang2024long}. In this case, we find that the model does not exhibit a complete uncertainty collapse -- even for very short random units of 5-9 residues repeated at a multiplicity of 32x (Figure~\ref{fig:Mirage}:D). We do however observe that the pseudo-perplexity declines steadily with increasing multiplicity (Figure~\ref{fig:Mirage}:D). This decline is most pronounced for the shortest repeating units, and gets weaker as the size of the repeating unit increases (Figure~\ref{fig:Mirage}:D).

\subsection{In-context retrieval can override learned priors}

The collapse in the pseudo-perplexity of doubled random sequences shows that transformer-based masked language models can perform a distinctive kind of in-context retrieval. For random sequences, this can only happen if the model can look up the identity of the residue at the equivalent position in the second copy of the sequence. In doing so, they effectively use one copy as a reference to completely resolve the uncertainty in the other. We will now explore how this phenomenology extends to protein domains, and examine related facets of the retrieval behavior for the transformer-based model ESM2 (650M). We will use our dataset of parsed domains that have an ESM2 OFS pseudo-perplexity~\citep{kantroo2024pseudo} score greater than five for our experiments -- the high pseudo-perplexity threshold is meant to filter out sequences that already contain repeated motifs.

Consider the representative protein domain in Figure~\ref{fig:Override}:A, that is masked at position 13. The model's prediction reveals a preference for amino acids with large hydrophobic side-chains like Phenylalanine (F), Tyrosine (Y) and Tryptophan (W), which aligns with the actual identity of the residue at this position: Tryptophan (W). Adding a second copy of the domain to this sequence induces a dramatic reduction in the uncertainty of the model -- it becomes absolutely confident that Tryptophan (W) is the only possible residue that can be placed here (Figure~\ref{fig:Override}:B). We now place an additional mask in the second copy of the domain at the equivalent position and find that the uncertainty in the prediction of the model returns, with a slight increase in the probability mass for other residues (Figure~\ref{fig:Override}:C). To rule out the possibility that the return of the uncertainty is purely a consequence of masking one additional position, as opposed to masking the equivalent position in particular, we now place a mask at a non-equivalent position in the second copy of the domain. In this case, the model maintains its conviction that Tryptophan (W) is the only viable residue for this position (Figure~\ref{fig:Override}:D). 

We scale this experiment to several thousand protein domains and measure the uncertainty in the model's predictions by the entropy of the probability vector. This recapitulates the findings laid out for the representative domain -- masking the equivalent position in the second copy is what disrupts the model's confidence (Figure~\ref{fig:Override}:E). This demonstrates that the model is indeed looking up the identity of the residue in the second copy of the protein domain in making its prediction. We note that this is spiritually similar to how induction heads work in autoregressive transformers~\citep{reddy2023mechanistic, olsson2022context, elhage2021mathematical}, although the underlying details of the circuits in bidirectional networks may differ. 

\begin{figure}
    \centering
    \includegraphics[width=1\textwidth]{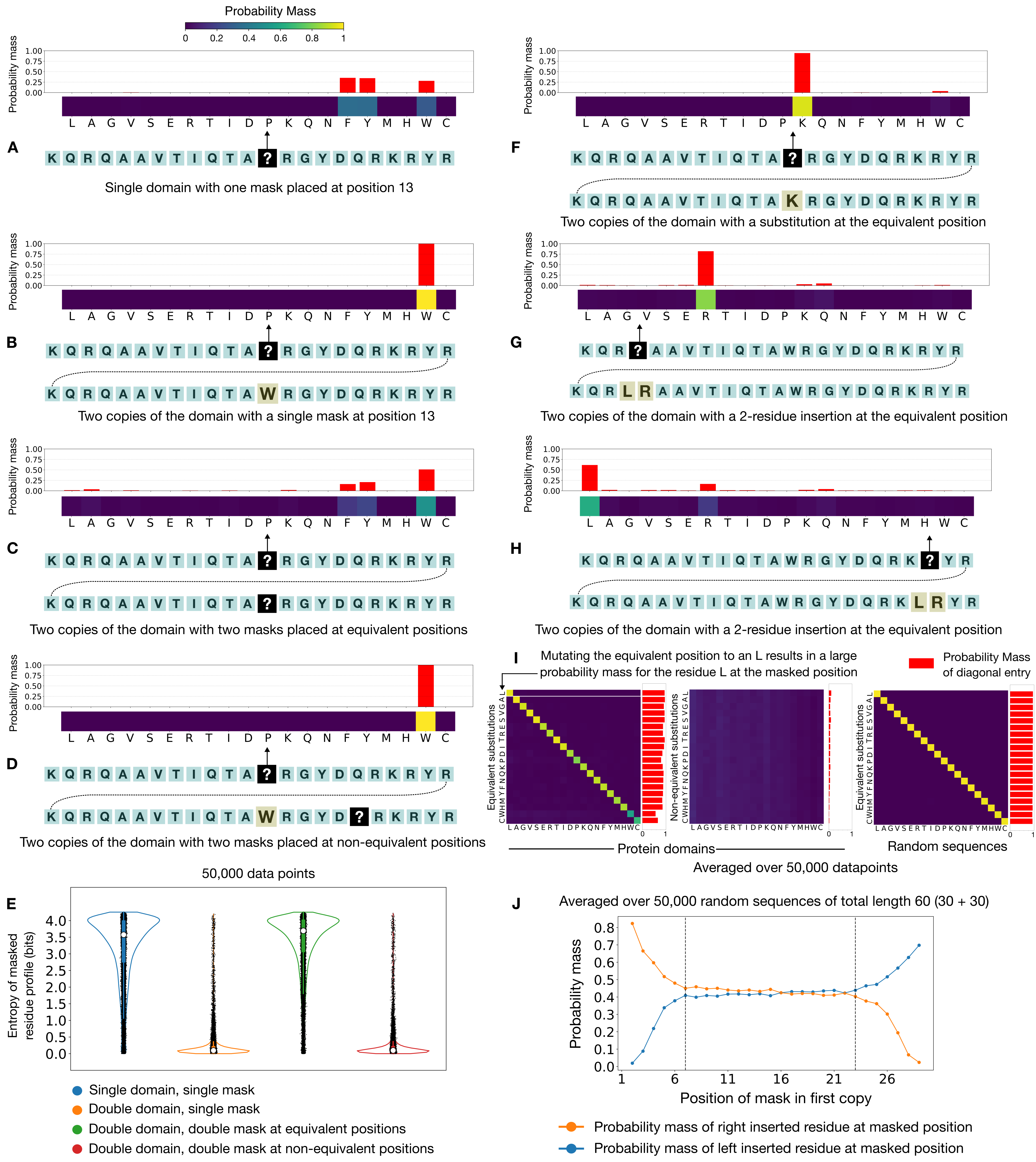}
    \caption{\small \textbf{In-context retrieval can override learned priors} 
                            (A) A depiction of a Calmodulin-binding motif (\texttt{Q622K8:757-779}) that is masked at position 13. The model's prediction for the masked position indicates a preference for residues with large hydrophobic side-chains. This aligns with the identity of the actual residue in the position -- W.
                            (B) Adding the second domain dramatically reduces the uncertainty of the model. The model is confident that W is the only sensible option for the masked position.
                            (C) Adding a mask at the equivalent position in the second copy of the domain brings back the uncertainty in the prediction of the model.
                            (D) Adding a mask at a non-equivalent position does not affect the model's confidence.
                            (E) Scaling the double-masking experiment to several thousand domains establishes that masking the equivalent position is what causes the return in the uncertainty of the model, as seen by the entropy distribution of the predictions.}
    \label{fig:Override}
\end{figure}

\begin{figure}
    \ContinuedFloat
    \caption[]{\small       (F) Changing the residue at the equivalent position to K                          also changes the model's prediction correspondingly.                             The fact that the model is otherwise aware that this                             residue does not fit in the given context does not                               deter it from making this prediction.
                            (G) The mask is placed at position 4 in the first copy of the domain. There are two equivalent positions in the second copy that the model can use for retrieval. In this case, the model prefers the residue on the right -- R.
                            (H) The mask is placed at position 21 in the first copy of the domain. There are two equivalent positions that the model can use for retrieval. In this case, the model prefers the residue on the left -- L.
                            (I) We scale the setup described in F to thousands of protein domains (leftmost panel) as well as randomly generated sequences (rightmost panel). Each row of the matrix is the averaged probability vector of the masked position when the equivalent position in the second copy is substituted by the amino acid corresponding to that row. We find that most of the probability mass is concentrated in the diagonal entries of the profile matrix for both protein domains (leftmost panel) and random sequences (rightmost panel). This means that the model's prediction for the masked position skews towards whatever residue is present at the equivalent position in the second copy. For natural protein domains, the probability mass for W and C is markedly lower than other residues (leftmost panel -- bottom two rows with dull green diagonal entries), indicating that changing the equivalent residue to W or C does not induce as strong of a flipping response in the model's prediction for the masked position. We do not observe this behavior for randomly generated sequences (rightmost panel -- all diagonal entries are bright yellow).
                            (J) A plot to show the model's contra-lateral preference for the retrieval position for the setup described in G and H: the model prefers the residue on the right for the retrieval operation when the mask is placed towards the left end of the sequence, and it prefers the residue on the left when the mask is placed on the right end. }
\end{figure}

The natural next question is: what if the retrieved residue from the second domain clashes with the model's assessment of the masked position?~\citep{wei2023larger} To pose this question, we replace the Tryptophan (W) at position 13 in the second copy of the domain with Lysine (K). Given that the model prefers residues with large hydrophobic side-chains like Tryptophan (W) in this position (Figure~\ref{fig:Override}:A), replacing it  with Lysine (K) is a particularly egregious choice because of its charged side-chain. With this change, the model flips to believing that Lysine (K) is the only valid option for this position (Figure~\ref{fig:Override}:F). This shows that at least in this case, the model resolves the conflict between its learned priors and the retrieved context by letting go of what it knows to be true and sticking with what it sees in the context. We repeat this experiment for several thousand domains in which we mutate the equivalent position in the second copy to a random residue and note the change in the model's prediction for the masked position. This experiment confirms that the model's predictions skew towards whatever residue appears at the equivalent position in the second copy (Figure~\ref{fig:Override}:I). This is in line with observations made in flipped-label setups in natural language where models have been shown to override learned semantic priors when presented with contradictory in-context exemplars~\citep{wei2023larger}.

Notice that the probability mass for two residues, Tryptophan (W) and Cysteine (C), is significantly lower in relation to other residues -- changing the equivalent position in the second copy of the domain to a Tryptophan (W) or Cysteine (C) does not yield as strong of a flipping response as compared to other residues (Figure~\ref{fig:Override}:I). Perhaps not coincidentally so, Tryptophan (W) and Cysteine~(C) happen to be the rarest amino acids in proteins~\citep{krick2014amino}. Repeating this experiment for randomly generated sequences reveals that this weaker response for Tryptophan (W) and Cysteine (C) does not persist in this case (Figure~\ref{fig:Override}:I). We interpret this to mean that even though learned priors can be overridden by in-context retrieval, they can still exert some influence on the model's predictions in this setup.

We next examine the situation when there is an ambiguity in how the model should perform the retrieval operation. We introduce this ambiguity by creating two retrieval sources that are presumably equally valid -- we remove the original residue at the equivalent position in the second domain and introduce a two-residue insertion (Figure~\ref{fig:Override}:G and H). In Figure~\ref{fig:Override}:G, we mask the representative domain at position 4 and present the model with two options that it can peek at to make its prediction. The model strongly prefers the residue on the right for the retrieval operation in this case. If we repeat this experiment for position 21 instead, we find that the model now prefers the residue on the left (Figure~\ref{fig:Override}:H). In view of these observations, we hypothesized that the  model's preference could depend on the location of the mask in the sequence: the model picks the residue on the right when the mask is placed at the left end of the sequence, and picks the residue on the left when it is placed on the right end. Experiments with randomly generated sequences of size 30 reveal that this is indeed the case -- the model exhibits a contra-lateral preference for the retrieval operation in this setup. We note that this behavior only manifests at the extreme ends of the sequence (Figure~\ref{fig:Override}:J). We hypothesize that this quirk may be traced to the model choosing to side with the longer contiguous repeating unit.

\subsection{In-context learning extends well beyond perfect repeats}

The low pseudo-perplexity scores of natural proteins with repeated motifs (Figure \ref{fig:schematic}) suggests that ESM2 (650M) can perform the retrieval operation even for imperfectly repeated sequences. We explicitly test this by concatenating a natural protein domain with progressively divergent copies of the sequence. These imperfect copies are constructed by introducing a fixed proportion of random mutations in the sequence -- substitutions, insertions and deletions are all admissible changes (Figure~\ref{fig:Beyond}:A). Since we have established that the low pseudo-perplexity scores of repeated sequences are an intrinsic feature of the model's behavior, and not a quirk of the One Fell Swoop method, we use OFS pseudo-perplexity to score the sequences. We find that the local OFS pseudo-perplexity of the mutated sequence is significantly lower when it appears alongside the natural copy, as compared to when it appears in isolation. This is evident from their respective distributions (Figure~\ref{fig:Beyond}:B). This shows that the model can retrieve information from the provided context even when half of the positions in the other copy of the sequence are different. 

One possible hypothesis for how this works is that pockets of perfectly repeated chunks can exist inside imperfectly repeated sequences. Once identified, these segments can be used as a reference for the retrieval operation. We probe this capability by performing a needle-in-a-haystack test~\citep{liu2024lost}. In this test, two identical copies of a random sequence -- the needles, are separated by a long stretch of an unrelated randomly generated sequence -- the haystack (Figure~\ref{fig:Beyond}:C). Since the needles are randomly generated, a low pseudo-perplexity score for one needle indicates that the model is using the needle placed at the other end of the sequence for retrieval. We vary the size of the needles and the haystack to characterize the limits of the retrieval behavior. The results indicate that the model can reliably perform the retrieval operation for needles as small as 10 residues even when the intervening haystacks span hundreds of residues (Figure~\ref{fig:Beyond}:D). In addition, we see that the transformer-based model ESM2 (650M) has an operational memory of approximately 1000 residues -- it does not recognize larger repeating units (Figure~\ref{fig:Beyond}:D).

Another possibility is that the retrieval behavior can be triggered even in the absence of contiguous matching segments. We test this hypothesis by creating randomly generated sequence pairs that share significant global similarity but contain no perfectly matching chunks: a one-skip sequence pair has matching residues in alternating positions (Figure~\ref{fig:Beyond}:E). We find that when presented with such sequences, the model behaves in a way as if there are two repeated motifs within the sequence. In Figure \ref{fig:Beyond}:E, we see that the prediction of the model favors residues in the equivalent positions for four contiguous masked positions. We then average over the one-at-a-time masked profile of thousands of such one-skip sequences calculated via OFS~\citep{kantroo2024pseudo}, and find that residues in the equivalent positions are indeed assigned a high probability mass by the model throughout the sequence (Figure \ref{fig:Beyond}:F). The probability profile of the correct residue however follows a spiking pattern that matches the construction of the sequences (Figure \ref{fig:Beyond}:F). These observations imply that the model can locate the equivalent retrieval source within these sequences to inform its predictions. If it exclusively relied on identifying contiguous matching segments, then the probability profile would have been similar to what we observe for unrelated random sequence pairs (Figure~\ref{fig:Beyond}:F). This suggests that the model can build a global view of the sequence through something akin to an internalized sequence alignment algorithm, rather than solely relying on exactly matching substrings. 

\begin{figure}[]
    \centering
    \includegraphics[width=1\textwidth]{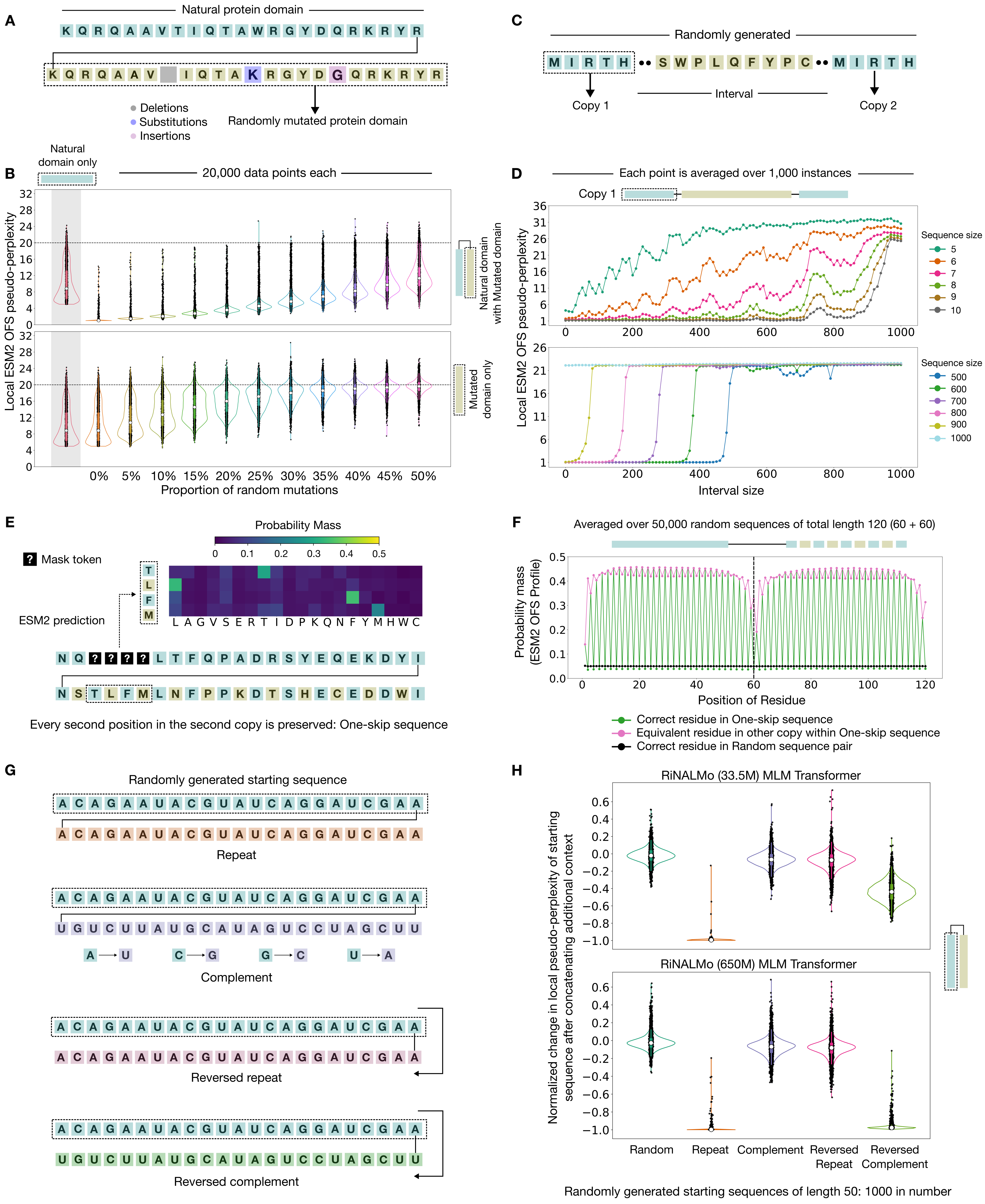}
    \caption{\small \textbf{In-context learning extends well beyond perfect repeats} 
                            (A) A representation of imperfectly repeated sequences. Mutations are added in the appended copy of the protein domain to assess if the in-context learning effect persists for imperfect repeats.
                            (B) The local ESM2 OFS pseudo-perplexity of the mutated protein domain is significantly lower when it appears alongside the natural domain as compared to when it appears in isolation.
                             }
    \label{fig:Beyond}
\end{figure}

\begin{figure}
    \ContinuedFloat
    \caption[]{\small        (C) A depiction of a sequence used in the needle in a haystack experiment with the needles (blue) at the two ends of the sequence and the haystack (yellow) in the middle.
                            (D) We track the local ESM2 OFS pseudo-perplexity of the needle placed at the beginning of the sequence. Needles as small as 10 residues can be recognized across haystacks spanning hundreds of residues. ESM2 cannot recognize repeats that are larger than about 1000 residues.
                            (E) One-skip sequences match in alternating positions as depicted in the figure -- the yellow residues in the second copy of the sequence do not match the blue residues in the first copy. Placing four contiguous masks in the sequence reveals that the model behaves in a way as if there are two repeats. It assigns a high probability mass for residues that are present at the equivalent position in the second copy.
                            (F) We determine the probability mass of the equivalent and correct residues in the sequence for the skip sequence context and compare it against two randomly generated sequences that have been concatenated. We find that the model assigns a high probability mass to the equivalent residue for the position in skip-sequences. The spikes in the probability mass of the correct residues match the construction of the sequence. This indicates that the model can identify the equivalent retrieval source within the skip-context to predict the identity of the residue at the masked position. The model could not have done so if it exclusively relied on contiguous matching segments.
                            (G) We use the transformer-based RNA language models RiNALMo 33.5M and 650M, and assess how appending additional context affects the pseudo-perplexity score of a randomly generated RNA sequence. We use repeats, complements, reversed repeats, and reversed complements of the original sequence as additional context. As a reference point, we use an unrelated random sequence as additional context.
                            (H) Repeats induce an uncertainty collapse in both the 33.5M and the 650M variants of the model. Hairpin-like motifs -- reversed complements, induce an uncertainty collapse in the 650M model, and only a partial reduction in uncertainty for the 33.5M variant of the model.}
\end{figure}

The ability to detect exact and imperfect repeats could arise from an enrichment of repeated sequences in the training data, from a quirk that stems from the model's architecture, or from a combination of these factors.  
Since the ability to learn in context is known to be sensitive to the statistical features of the training dataset~\citep{chan2022data}, we wanted to explore if the model could detect sequences that have undergone biologically meaningful transformations. Specifically, RNA sequences are known to contain secondary structure motifs called hairpins~\citep{svoboda2006hairpin} that contain stretches of nucleotides which are reversed and complemented versions of the preceding sequence (e.g. AUCCG is the reversed complement of CGGAU). These stretches are physically meaningful since they can pair with one another. Their role as a secondary structure motif means that they are enriched in biological RNA sequences.  Importantly, while reversed complements are biologically meaningful, complemented stretches (e.g. AUCCG is the complement of UAGGC) and reversed sequences (e.g. AUCCG is GCCUA reversed) on their own imply no particular physical interactions, and are not biologically meaningful. We hypothesized that transformer-based RNA language models, RiNALMo (33.5 M) and RiNALMo (650 M), may recognize the reversed complement motif for in-context retrieval. We test this by examining how appending additional context to a randomly generated RNA sequence affects its pseudo-perplexity -- a significant drop should be indicative of in-context learning. We perform this experiment by using five kinds of sequences as additional context: an unrelated random sequence, a direct repeat, the complementary sequence, a reversed repeat, and finally the complement of the reversed copy of the original random sequence -- resulting in a hairpin-like motif (Figure~\ref{fig:Beyond}:G). 

We find that much like proteins, repeats induce an uncertainty collapse in both the 33.5M and the 650M variants of the masked RNA language model (Figure~\ref{fig:Beyond}:H). However, the hairpin-like motifs induce a complete uncertainty collapse in the 650M variant, and only a partial reduction in uncertainty for the 33.5M variant of the model (Figure~\ref{fig:Beyond}:H). The larger 650M model thus behaves differently even when both the model variants are presented with the same underlying pattern of hairpin-like motifs. More notably, addition of reversed repeats and complemented sequences to the context does not trigger in-context learning for either variant of the model (Figure~\ref{fig:Beyond}:H). Reversed complements are biologically meaningful sequences that form secondary structure motifs, and are hence well represented in the training dataset of the model. Sequences that have only been reversed or complemented however are bound to be far less prevalent since they are not relevant to biology. This reinforces the idea that in-context learning capabilities sensitively depend on the statistical features of the dataset -- only features and patterns that the model has extensively seen during training trigger the retrieval behavior in this case. This in turn demonstrates that composite patterns that the model has learned to recognize are not learned in a decomposable fashion -- although the model can recognize the reversed complement of a sequence, it is unable to perform the retrieval operation for sequences that have only been reversed or complemented.

\subsection{Repetition can deteriorate the quality of embeddings}

\begin{figure}[!]
    \centering
    \includegraphics[width=1\textwidth]{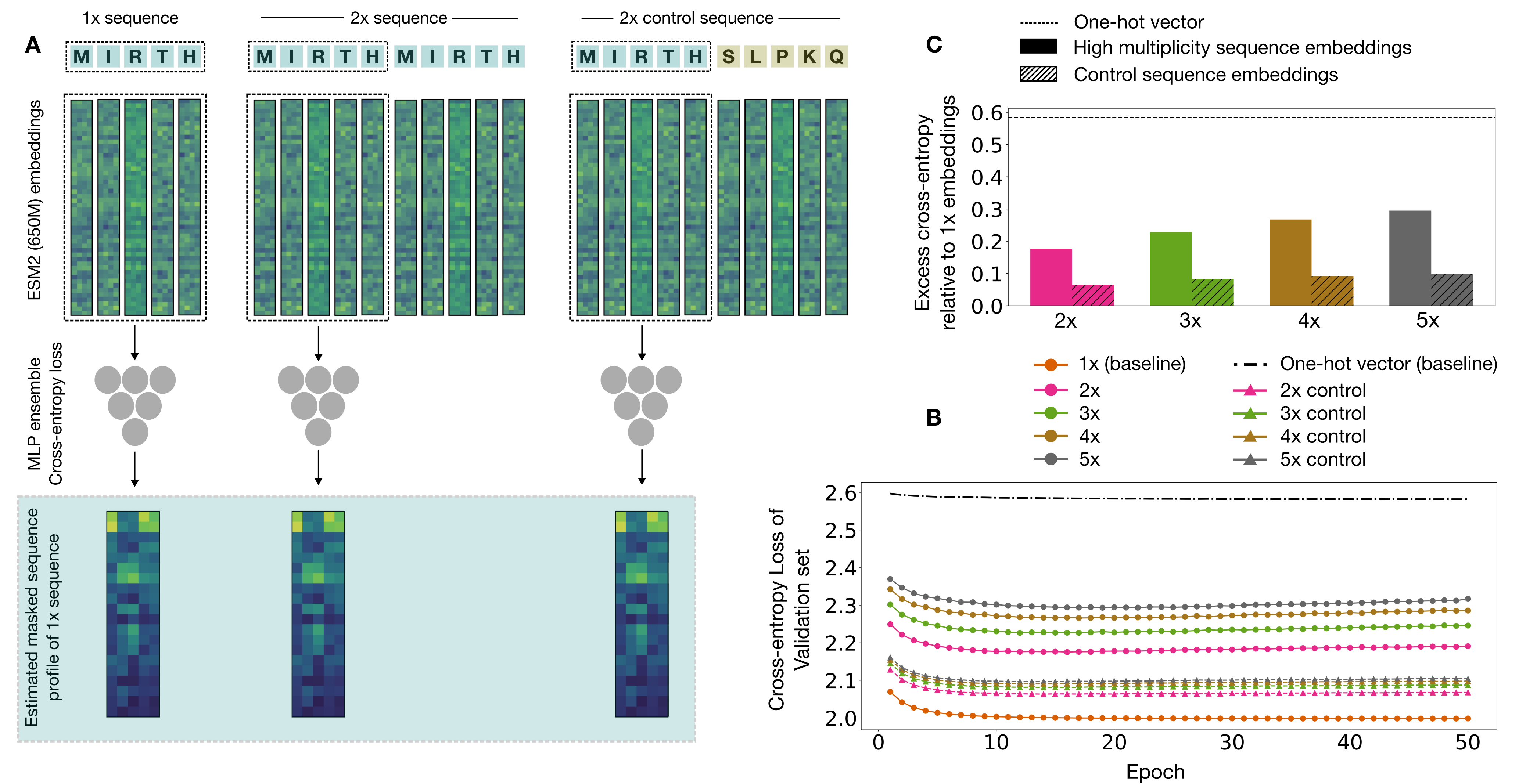}
    \caption{\small \textbf{Repetition can deteriorate the quality of embeddings} 
                            (A) A depiction of the One Fell Swoop regression task as it is used in this setting. The embedding vectors are used to regress on the one-at-a-time masked profile of the sequence, and the loss achieved by the model is used as a proxy for the information content of the embeddings -- lower the loss, higher the information content. We use 1x sequence embeddings and one-hot vectors as the baselines, and compare the higher multiplicity versions of the sequences against their respective control variants.
                            (B) The validation loss curves for the respective embedding vectors averaged over 5 training runs with different training-validation splits. 
                            (C) The validation loss achieved by the high multiplicity embeddings is higher than the embeddings from the control sequences for this regression task. It continues to increase for higher multiplicity embeddings. The detrimental impact of repetition is significantly stronger than appending randomly generated sequences to protein sequences in this case.}
    \label{fig:Embedding}
\end{figure}

Masked language models are primarily used to transform an input sequence into a set of high dimensional vectors called the embedding of the sequence. These embedding vectors are informative representations of the input data, and can serve as the foundation for a host of downstream tasks~\citep{kilgore2025protein, wang2023netgo}. We will now assess how repetition impacts the information content of residue-level embeddings of protein sequences as generated by the masked language model ESM2 (650M). We do so by assessing how well a given set of embedding vectors does on the One Fell Swoop regression task~\citep{kantroo2024pseudo}. This involves using the embeddings of the residues in a sequence to predict its one-at-a-time masked profile (Figure \ref{fig:Embedding}:A). The loss achieved by the model on this regression task is our proxy for the information content of the embeddings: lower loss can be attributed to higher information content of the embedding vectors.

We first generate the embedding vectors for a large corpus of natural proteins (see Methods for details) along with their higher multiplicity variants (2x to 5x). We repeat this calculation for control sequences that are the same length as the higher multiplicity variants but only contain a single copy of the protein sequence. The natural protein sequence is placed in the beginning of the control sequence and the rest of it is randomly generated (Figure~\ref{fig:Embedding}:A). We compare the performance of embeddings from the first repeating unit from the higher multiplicity sequence variants against the corresponding unit in the higher multiplicity control sequences (dotted outline in Figure~\ref{fig:Embedding}:A). The 1x sequence embeddings set the baseline for how well a model does in the standard One Fell Swoop setting for this training dataset. We also use one-hot vector representations as a baseline since they only contain information about the identity of the residue at a given position and are devoid of context-sensitive information. This setup allows us to compare how repetition affects the quality of residue embeddings relative to appending a randomly generated nonsensical sequence to a natural protein.

We feed the generated embedding vectors of the respective groups to an ensemble of MLP models. These models are trained to predict the one-at-a-time masked profile of a sequence by minimizing the cross-entropy loss between the prediction and the target distribution (Figure~\ref{fig:Embedding}:A). We train a separate ensemble of MLP models for each multiplicity and control group of sequences -- 10 models in total for a given training-validation data split, 80-20 in this case. We find that models trained on higher multiplicity embeddings achieve a higher validation loss as compared to models trained on the embeddings of the respective control sequences on this task. We also note that the validation loss increases with higher sequence multiplicity (Figure~\ref{fig:Embedding}:B and Figure~\ref{fig:Embedding}:C). Repetition thus has a detrimental impact on the information content of the embeddings generated by ESM2 for this regression task. This effect is significantly stronger than the dilutive effect of adding long randomly generated sequences to natural proteins. 

\section{Discussion}

In-context learning is a striking feature of large language models that allows them to learn from the presented context on the fly. Although this is eminently useful for generative models trained on natural language, pattern recognition of surface-level features can have undesirable effects in the setting of biological sequences. In particular, it distorts the relationship between fitness and likelihood scores by making otherwise nonsensical sequences seem natural. More importantly however, it can disallow a model from inferring reliable mutation profiles for sequences that are truly functional -- protein families with repeated domains for instance~\citep{andrade2001protein}. Given their preponderance and physiological relevance, this counts as an important failure mode for models that are exclusively trained on the sequence modality. However, this behavior need not be exclusive to the sequence domain -- a model trained on biomolecular structures could perhaps use symmetric subunits within a structure to perform in-context retrieval. More generally, a model could exploit all kinds of symmetries within a training dataset and learn recurring patterns -- as has also been found in the natural language setting~\citep{chen2024parallel}.

We find that the ability to learn from the provided context significantly varies across different architectures. The transformer-based masked language model ESM2~\citep{lin2023evolutionary} exhibits a distinctive kind of in-context retrieval that manifests as an uncertainty collapse when presented with repeated sequences. ESM2 uses other copies of a repeated motif to completely resolve the uncertainty in its predictions for a given masked position. Consequently, the pseudo-perplexity score of such sequences is approximately one -- the lowest value that it can take. This is because pseudo-perplexity is defined in terms of the one-at-a-time masked profile of a sequence. The BiMamba-S based large context protein language model~\citep{wang2024long} on the other hand only exhibits a progressive decline in uncertainty as the multiplicity of the repeating motif is increased. Intriguingly, the convolutions-based protein language model CARP (640M)~\citep{yang2024convolutions} exhibits a transition between these two kinds of behaviors: it exhibits uncertainty collapse for repeating units that are shorter than 70 residues, and progressive decline for larger repeating units. This behavior makes CARP (640M) an excellent specimen to be dissected for interpretability purposes~\citep{rai2024practical}. It would also be prudent to monitor the dynamics of the emergence of this ability over training given the transient nature of in-context learning~\citep{singh2023transient}.

Larger models are known to be more adept at utilizing the provided context~\citep{brown2020language, elhage2021mathematical}. Consistent with this observation, we note that larger models induce larger distortions in the likelihood scores even when presented with the same pattern -- the 650M variant of RiNALMo~\citep{penic2024rinalmo} exhibits an uncertainty collapse for hairpin-like motifs, while the 33.5M variant only exhibits a partial reduction in its uncertainty. Notably, the skill to recognize hairpin-like motifs that contain a sequence along with its reversed complement has not been acquired in a decomposable manner -- neither of these models is able to perform in-context retrieval for sequences that have only been reversed or complemented. Although there is considerable interest in the skill composition capabilities of large language models~\citep{yu2023skill, zhao2024can}, this observation points towards the need for a systematic inquiry into their ability to recover simpler sub-skills from learned composite skills. Despite the data-dependence of the retrieval capability, unforeseen and hard to predict quirks can still emerge in these models as seen by the contra-lateral retrieval preference exhibited by ESM2. Additionally, we find that these models can implement sophisticated retrieval algorithms such as an internalized sequence alignment to identify appropriate retrieval sources within the context. 

More broadly, our results imply that a lower average perplexity or pseudo-perplexity score need not always translate to a deeper understanding of evolutionary or biophysical constraints. It could instead reflect the model's proficiency with pattern recognition when the underlying dataset is especially replete with degenerate motifs. This effect can become especially pernicious for larger models that could potentially also recognize non-obvious semantic patterns -- well beyond just surface-level features. In such cases, it is possible for in-context learning induced distortions to be misdiagnosed as memorization-related artifacts. This possibility presents an interesting conundrum because the ability to capture context-specific relationships is what allows these models to build informative representations in the first place. A natural next step in this line of questioning is to investigate if the known biases exhibited by these models are connected to this phenomenology.

\section{Methods}

We parsed protein sequences in the Swissprot database~\citep{uniprot2015uniprot} into their constituent domains using RPS-BLAST~\citep{marchler2004cd}. We filtered out domains that are shorter than 20 residues and longer than 1000 residues. We clustered the resulting set of sequences via mmseqs2~\citep{steinegger2017mmseqs2} using a minimum sequence identity of 0.3 and coverage mode 1. This yields a set of 76,361 protein domains. We use the ESM2 OFS pseudo-perplexity of this set of parsed domains for the plot in Figure 1. We sample 1000 domains from the set of 76,361 domains that are shorter than 400 residues for the experiments in Figure 2. We filter out domains that have an ESM2 OFS pseudo-perplexity less than 5 for the experiments in figure 3 and 4 -- this is to remove sequences that already contain repeated motifs. We did not use the first position of a sequence in our experiments because of the known methionine bias in this position. In figure 5, we use a subset of the sequences used to train the One Fell Swoop model~\citep{kantroo2024pseudo}. This subset consists of sequences within the original dataset that are shorter than 200 residues resulting in 37,176 protein sequences in total. We use an 80-20 training-validation split for 5 separate training runs. The validation loss curves in the plot are averaged over these 5 runs that use different splits.

\section{Acknowledgements}

We thank Michael Abbott and Casey Dunn for close reads and useful feedback on the manuscript. This work was supported by NIH R35GM138341 (BBM). We also thank the Yale Center for Research Computing, ITS Cloud, and AWS for guidance and use of AWS Cloud computing infrastructure through Cloud Credits for Research Program.

\bibliographystyle{ieeetr}
\bibliography{refs.bib}

\begin{thebibliography}{10}

\bibitem{lin2023evolutionary}
Z.~Lin, H.~Akin, R.~Rao, B.~Hie, Z.~Zhu, W.~Lu, N.~Smetanin, R.~Verkuil, O.~Kabeli, Y.~Shmueli, {\em et~al.}, ``Evolutionary-scale prediction of atomic-level protein structure with a language model,'' {\em Science}, vol.~379, no.~6637, pp.~1123--1130, 2023.

\bibitem{zhou2023dnabert2}
Z.~Zhou, Y.~Ji, W.~Li, P.~Dutta, R.~Davuluri, and H.~Liu, ``Dnabert-2: Efficient foundation model and benchmark for multi-species genome,'' 2023.

\bibitem{penic2024rinalmo}
R.~J. Peni{\'c}, T.~Vla{\v{s}}i{\'c}, R.~G. Huber, Y.~Wan, and M.~{\v{S}}iki{\'c}, ``Rinalmo: General-purpose rna language models can generalize well on structure prediction tasks,'' {\em arXiv preprint arXiv:2403.00043}, 2024.

\bibitem{theodoris2023transfer}
C.~V. Theodoris, L.~Xiao, A.~Chopra, M.~D. Chaffin, Z.~R. Al~Sayed, M.~C. Hill, H.~Mantineo, E.~M. Brydon, Z.~Zeng, X.~S. Liu, {\em et~al.}, ``Transfer learning enables predictions in network biology,'' {\em Nature}, vol.~618, no.~7965, pp.~616--624, 2023.

\bibitem{trauble2025multi}
F.~Tr{\"a}uble, L.~Stuart, A.~Georgiou, P.~Notin, A.~Mehrjou, R.~Schwessinger, M.~Chevalley, K.~Branson, B.~Sch{\"o}lkopf, C.~van Duijn, {\em et~al.}, ``Multi-megabase scale genome interpretation with genetic language models,'' {\em arXiv preprint arXiv:2501.07737}, 2025.

\bibitem{nijkamp2023progen2}
E.~Nijkamp, J.~A. Ruffolo, E.~N. Weinstein, N.~Naik, and A.~Madani, ``Progen2: exploring the boundaries of protein language models,'' {\em Cell systems}, vol.~14, no.~11, pp.~968--978, 2023.

\bibitem{nguyen2024sequence}
E.~Nguyen, M.~Poli, M.~G. Durrant, B.~Kang, D.~Katrekar, D.~B. Li, L.~J. Bartie, A.~W. Thomas, S.~H. King, G.~Brixi, {\em et~al.}, ``Sequence modeling and design from molecular to genome scale with evo,'' {\em Science}, vol.~386, no.~6723, p.~eado9336, 2024.

\bibitem{shulgina2024rna}
Y.~Shulgina, M.~I. Trinidad, C.~J. Langeberg, H.~Nisonoff, S.~Chithrananda, P.~Skopintsev, A.~J. Nissley, J.~Patel, R.~S. Boger, H.~Shi, {\em et~al.}, ``Rna language models predict mutations that improve rna function,'' {\em Nature Communications}, vol.~15, no.~1, pp.~1--17, 2024.

\bibitem{ruffolo2024designing}
J.~A. Ruffolo and A.~Madani, ``Designing proteins with language models,'' {\em Nature Biotechnology}, vol.~42, no.~2, pp.~200--202, 2024.

\bibitem{notin2024proteingym}
P.~Notin, A.~Kollasch, D.~Ritter, L.~Van~Niekerk, S.~Paul, H.~Spinner, N.~Rollins, A.~Shaw, R.~Orenbuch, R.~Weitzman, {\em et~al.}, ``Proteingym: large-scale benchmarks for protein fitness prediction and design,'' {\em Advances in Neural Information Processing Systems}, vol.~36, 2024.

\bibitem{benegas2025benchmarking}
G.~Benegas, G.~Eraslan, and Y.~S. Song, ``Benchmarking dna sequence models for causal regulatory variant prediction in human genetics,'' {\em bioRxiv}, pp.~2025--02, 2025.

\bibitem{brixi2025genome}
G.~Brixi, M.~G. Durrant, J.~Ku, M.~Poli, G.~Brockman, D.~Chang, G.~A. Gonzalez, S.~H. King, D.~B. Li, A.~T. Merchant, {\em et~al.}, ``Genome modeling and design across all domains of life with evo 2,'' {\em bioRxiv}, pp.~2025--02, 2025.

\bibitem{ding2024protein}
F.~Ding and J.~Steinhardt, ``Protein language models are biased by unequal sequence sampling across the tree of life,'' {\em BioRxiv}, pp.~2024--03, 2024.

\bibitem{shaw2023removing}
A.~Y. Shaw, H.~B. Spinner, S.~Gurev, J.-E. Shin, N.~Rollins, and D.~S. Marks, ``Removing bias in sequence models of protein fitness,'' {\em bioRxiv}, pp.~2023--09, 2023.

\bibitem{gordon2024protein}
C.~Gordon, A.~X. Lu, and P.~Abbeel, ``Protein language model fitness is a matter of preference,'' {\em bioRxiv}, pp.~2024--10, 2024.

\bibitem{salazar2019masked}
J.~Salazar, D.~Liang, T.~Q. Nguyen, and K.~Kirchhoff, ``Masked language model scoring,'' {\em arXiv preprint arXiv:1910.14659}, 2019.

\bibitem{marchler2004cd}
A.~Marchler-Bauer and S.~H. Bryant, ``Cd-search: protein domain annotations on the fly,'' {\em Nucleic acids research}, vol.~32, no.~suppl\_2, pp.~W327--W331, 2004.

\bibitem{lu2020cdd}
S.~Lu, J.~Wang, F.~Chitsaz, M.~K. Derbyshire, R.~C. Geer, N.~R. Gonzales, M.~Gwadz, D.~I. Hurwitz, G.~H. Marchler, J.~S. Song, {\em et~al.}, ``Cdd/sparcle: the conserved domain database in 2020,'' {\em Nucleic acids research}, vol.~48, no.~D1, pp.~D265--D268, 2020.

\bibitem{kantroo2024pseudo}
P.~Kantroo, G.~Wagner, and B.~Machta, ``Pseudo-perplexity in one fell swoop for protein fitness estimation,'' {\em bioRxiv}, pp.~2024--07, 2024.

\bibitem{reddy2023mechanistic}
G.~Reddy, ``The mechanistic basis of data dependence and abrupt learning in an in-context classification task,'' {\em arXiv preprint arXiv:2312.03002}, 2023.

\bibitem{chan2022data}
S.~Chan, A.~Santoro, A.~Lampinen, J.~Wang, A.~Singh, P.~Richemond, J.~McClelland, and F.~Hill, ``Data distributional properties drive emergent in-context learning in transformers,'' {\em Advances in neural information processing systems}, vol.~35, pp.~18878--18891, 2022.

\bibitem{singh2023transient}
A.~Singh, S.~Chan, T.~Moskovitz, E.~Grant, A.~Saxe, and F.~Hill, ``The transient nature of emergent in-context learning in transformers,'' {\em Advances in Neural Information Processing Systems}, vol.~36, pp.~27801--27819, 2023.

\bibitem{elhage2021mathematical}
N.~Elhage, N.~Nanda, C.~Olsson, T.~Henighan, N.~Joseph, B.~Mann, A.~Askell, Y.~Bai, A.~Chen, T.~Conerly, {\em et~al.}, ``A mathematical framework for transformer circuits,'' {\em Transformer Circuits Thread}, vol.~1, no.~1, p.~12, 2021.

\bibitem{olsson2022context}
C.~Olsson, N.~Elhage, N.~Nanda, N.~Joseph, N.~DasSarma, T.~Henighan, B.~Mann, A.~Askell, Y.~Bai, A.~Chen, {\em et~al.}, ``In-context learning and induction heads,'' {\em arXiv preprint arXiv:2209.11895}, 2022.

\bibitem{brown2020language}
T.~Brown, B.~Mann, N.~Ryder, M.~Subbiah, J.~D. Kaplan, P.~Dhariwal, A.~Neelakantan, P.~Shyam, G.~Sastry, A.~Askell, {\em et~al.}, ``Language models are few-shot learners,'' {\em Advances in neural information processing systems}, vol.~33, pp.~1877--1901, 2020.

\bibitem{akyurek2022learning}
E.~Aky{\"u}rek, D.~Schuurmans, J.~Andreas, T.~Ma, and D.~Zhou, ``What learning algorithm is in-context learning? investigations with linear models,'' {\em arXiv preprint arXiv:2211.15661}, 2022.

\bibitem{von2023transformers}
J.~Von~Oswald, E.~Niklasson, E.~Randazzo, J.~Sacramento, A.~Mordvintsev, A.~Zhmoginov, and M.~Vladymyrov, ``Transformers learn in-context by gradient descent,'' in {\em International Conference on Machine Learning}, pp.~35151--35174, PMLR, 2023.

\bibitem{park2024competition}
C.~F. Park, E.~S. Lubana, I.~Pres, and H.~Tanaka, ``Competition dynamics shape algorithmic phases of in-context learning,'' {\em arXiv preprint arXiv:2412.01003}, 2024.

\bibitem{vaswani2017attention}
A.~Vaswani, N.~Shazeer, N.~Parmar, J.~Uszkoreit, L.~Jones, A.~N. Gomez, {\L}.~Kaiser, and I.~Polosukhin, ``Attention is all you need,'' {\em Advances in neural information processing systems}, vol.~30, 2017.

\bibitem{devlin2018bert}
J.~Devlin, M.-W. Chang, K.~Lee, and K.~Toutanova, ``Bert: Pre-training of deep bidirectional transformers for language understanding,'' {\em arXiv preprint arXiv:1810.04805}, 2018.

\bibitem{radford2019language}
A.~Radford, J.~Wu, R.~Child, D.~Luan, D.~Amodei, I.~Sutskever, {\em et~al.}, ``Language models are unsupervised multitask learners,'' {\em OpenAI blog}, vol.~1, no.~8, p.~9, 2019.

\bibitem{yang2024convolutions}
K.~K. Yang, N.~Fusi, and A.~X. Lu, ``Convolutions are competitive with transformers for protein sequence pretraining,'' {\em Cell Systems}, vol.~15, no.~3, pp.~286--294, 2024.

\bibitem{wang2024long}
Y.~Wang, Z.~Wang, G.~Sadeh, L.~Zancato, A.~Achille, G.~Karypis, and H.~Rangwala, ``Long-context protein language model,'' {\em bioRxiv}, pp.~2024--10, 2024.

\bibitem{wei2023larger}
J.~Wei, J.~Wei, Y.~Tay, D.~Tran, A.~Webson, Y.~Lu, X.~Chen, H.~Liu, D.~Huang, D.~Zhou, {\em et~al.}, ``Larger language models do in-context learning differently,'' {\em arXiv preprint arXiv:2303.03846}, 2023.

\bibitem{krick2014amino}
T.~Krick, N.~Verstraete, L.~G. Alonso, D.~A. Shub, D.~U. Ferreiro, M.~Shub, and I.~E. S{\'a}nchez, ``Amino acid metabolism conflicts with protein diversity,'' {\em Molecular biology and evolution}, vol.~31, no.~11, pp.~2905--2912, 2014.

\bibitem{liu2024lost}
N.~F. Liu, K.~Lin, J.~Hewitt, A.~Paranjape, M.~Bevilacqua, F.~Petroni, and P.~Liang, ``Lost in the middle: How language models use long contexts,'' {\em Transactions of the Association for Computational Linguistics}, vol.~12, pp.~157--173, 2024.

\bibitem{svoboda2006hairpin}
P.~Svoboda and A.~D. Cara, ``Hairpin rna: a secondary structure of primary importance,'' {\em Cellular and Molecular Life Sciences CMLS}, vol.~63, pp.~901--908, 2006.

\bibitem{kilgore2025protein}
H.~R. Kilgore, I.~Chinn, P.~G. Mikhael, I.~Mitnikov, C.~Van~Dongen, G.~Zylberberg, L.~Afeyan, S.~F. Banani, S.~Wilson-Hawken, T.~I. Lee, {\em et~al.}, ``Protein codes promote selective subcellular compartmentalization,'' {\em Science}, p.~eadq2634, 2025.

\bibitem{wang2023netgo}
S.~Wang, R.~You, Y.~Liu, Y.~Xiong, and S.~Zhu, ``Netgo 3.0: protein language model improves large-scale functional annotations,'' {\em Genomics, Proteomics \& Bioinformatics}, vol.~21, no.~2, pp.~349--358, 2023.

\bibitem{andrade2001protein}
M.~A. Andrade, C.~Perez-Iratxeta, and C.~P. Ponting, ``Protein repeats: structures, functions, and evolution,'' {\em Journal of structural biology}, vol.~134, no.~2-3, pp.~117--131, 2001.

\bibitem{chen2024parallel}
Y.~Chen, C.~Zhao, Z.~Yu, K.~McKeown, and H.~He, ``Parallel structures in pre-training data yield in-context learning,'' {\em arXiv preprint arXiv:2402.12530}, 2024.

\bibitem{rai2024practical}
D.~Rai, Y.~Zhou, S.~Feng, A.~Saparov, and Z.~Yao, ``A practical review of mechanistic interpretability for transformer-based language models,'' {\em arXiv preprint arXiv:2407.02646}, 2024.

\bibitem{yu2023skill}
D.~Yu, S.~Kaur, A.~Gupta, J.~Brown-Cohen, A.~Goyal, and S.~Arora, ``Skill-mix: A flexible and expandable family of evaluations for ai models,'' {\em arXiv preprint arXiv:2310.17567}, 2023.

\bibitem{zhao2024can}
H.~Zhao, S.~Kaur, D.~Yu, A.~Goyal, and S.~Arora, ``Can models learn skill composition from examples?,'' {\em Advances in Neural Information Processing Systems}, vol.~37, pp.~102393--102427, 2024.

\bibitem{uniprot2015uniprot}
U.~Consortium, ``Uniprot: a hub for protein information,'' {\em Nucleic acids research}, vol.~43, no.~D1, pp.~D204--D212, 2015.

\bibitem{steinegger2017mmseqs2}
M.~Steinegger and J.~S{\"o}ding, ``Mmseqs2 enables sensitive protein sequence searching for the analysis of massive data sets,'' {\em Nature biotechnology}, vol.~35, no.~11, pp.~1026--1028, 2017.

\end{thebibliography}

\end{document}